\documentclass{article} 
\usepackage{iclr2025_conference,times}

\usepackage{amsmath,amsfonts,bm}

\def\eqref#1{equation~\ref{#1}}

\def\1{\bm{1}}

\DeclareMathAlphabet{\mathsfit}{\encodingdefault}{\sfdefault}{m}{sl}
\SetMathAlphabet{\mathsfit}{bold}{\encodingdefault}{\sfdefault}{bx}{n}

\usepackage{hyperref}
\usepackage{url}
\usepackage{graphicx}
\usepackage{cleveref}
\usepackage{booktabs}
\usepackage{subcaption}
\usepackage{multirow}
\usepackage{float}

\title{FLIP Reasoning Challenge}

\iclrfinalcopy

\author{Andreas Plesner \\
ETH Zurich\\
\texttt{aplesner@ethz.ch} \\
\And
Turlan Kuzhagaliyev \\
ETH Zurich 
\And
Roger Wattenhofer \\
ETH Zurich \\
\texttt{wattenhofer@ethz.ch}
}

\begin{document}

\maketitle

\begin{abstract}
    Over the past years, advances in artificial intelligence (AI) have demonstrated how AI can solve many perception and generation tasks, such as image classification and text writing, yet reasoning remains a challenge.
    This paper introduces the FLIP dataset, a benchmark for evaluating AI reasoning capabilities based on human verification tasks on the Idena blockchain. FLIP challenges present users with two orderings of 4 images, requiring them to identify the logically coherent one. By emphasizing sequential reasoning, visual storytelling, and common sense, FLIP provides a unique testbed for multimodal AI systems.
    
    Our experiments evaluate state-of-the-art models, leveraging both vision-language models (VLMs) and large language models (LLMs). Results reveal that even the best open-sourced and closed-sourced models achieve maximum accuracies of 75.5\% and 77.9\%, respectively, in zero-shot settings, compared to human performance of 95.3\%. Captioning models aid reasoning models by providing text descriptions of images, yielding better results than when using the raw images directly, 69.6\% vs. 75.2\% for \textsc{Gemini 1.5 Pro}. Combining the predictions from 15 models in an ensemble increases the accuracy to 85.2\%.
    These findings highlight the limitations of existing reasoning models and the need for robust multimodal benchmarks like FLIP. 
    The full codebase and dataset will be available at \url{https://github.com/aplesner/FLIP-Reasoning-Challenge}. 

\end{abstract}
\section{Introduction}

The rapid advancements in machine learning (ML) and artificial intelligence (AI) have led to impressive achievements across various domains, such as object detection and natural language processing. Despite the progress, reasoning—the ability to derive logical conclusions and establish relationships between abstract concepts—remains a significant challenge for AI. Reasoning is a hallmark of human intelligence but has proven difficult to evaluate and replicate in machines.

For many years, CAPTCHA systems such as Google reCAPTCHA and hCaptcha relied on the flaws of vision models to differentiate humans from bots. However, recent works demonstrate that machine algorithms now achieve near-perfect performance on these tasks, rendering traditional CAPTCHAs ineffective as measures of ``humanness.'' For instance, \citet{plesner2024breaking} showed that reCAPTCHAv2 could be solved by machines with a 100\% success rate, underlining the need for new benchmarks beyond recognition tasks.

Reasoning is a fundamental capability that AI models must master to perform tasks requiring comprehension, logical inference, and contextual understanding. Existing reasoning benchmarks, such as GSM8K \cite{mirzadeh2024gsmsymbolicunderstandinglimitationsmathematical}, CLEVR \cite{johnson2016clevrdiagnosticdatasetcompositional}, and MATH \cite{hendrycks2021measuringmathematicalproblemsolving}, provide valuable insights into AI reasoning but suffer from several limitations:
\begin{itemize}
\item \textbf{Data Contamination:} Many reasoning benchmarks inadvertently overlap with training data, leading to inflated performance metrics that do not reflect true reasoning capabilities.
\item \textbf{Narrow Scope:} These benchmarks often focus on specific domains, such as mathematical or visual reasoning, without capturing the multimodal and contextual complexity of real-world tasks.
\end{itemize}

\begin{figure}[t]
    \centering
    \begin{minipage}{0.3\textwidth}
        \centering
        \includegraphics[width=0.60\linewidth]{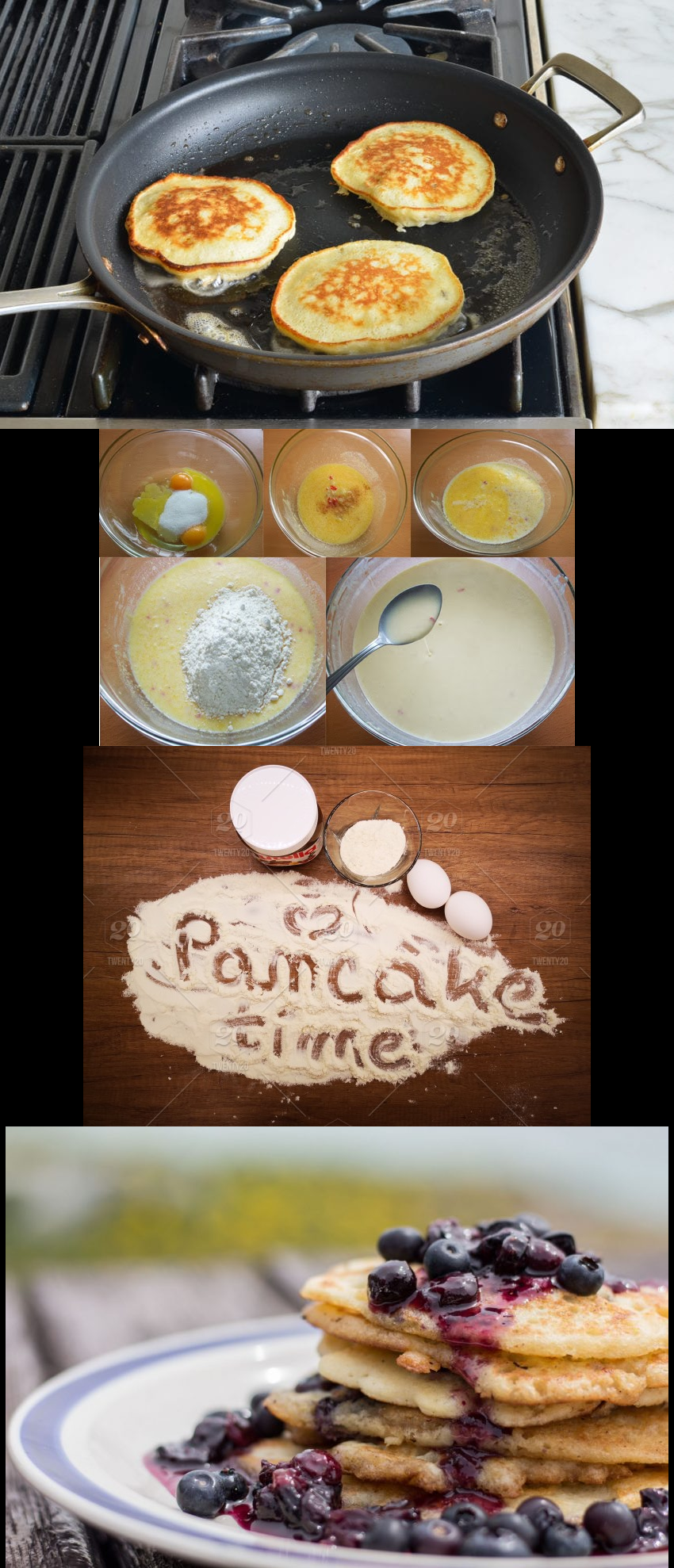}
        \caption*{(a) Left stack}
    \end{minipage}
    \hfill
    \begin{minipage}{0.3\textwidth}
        \centering
        \includegraphics[width=0.60\linewidth]{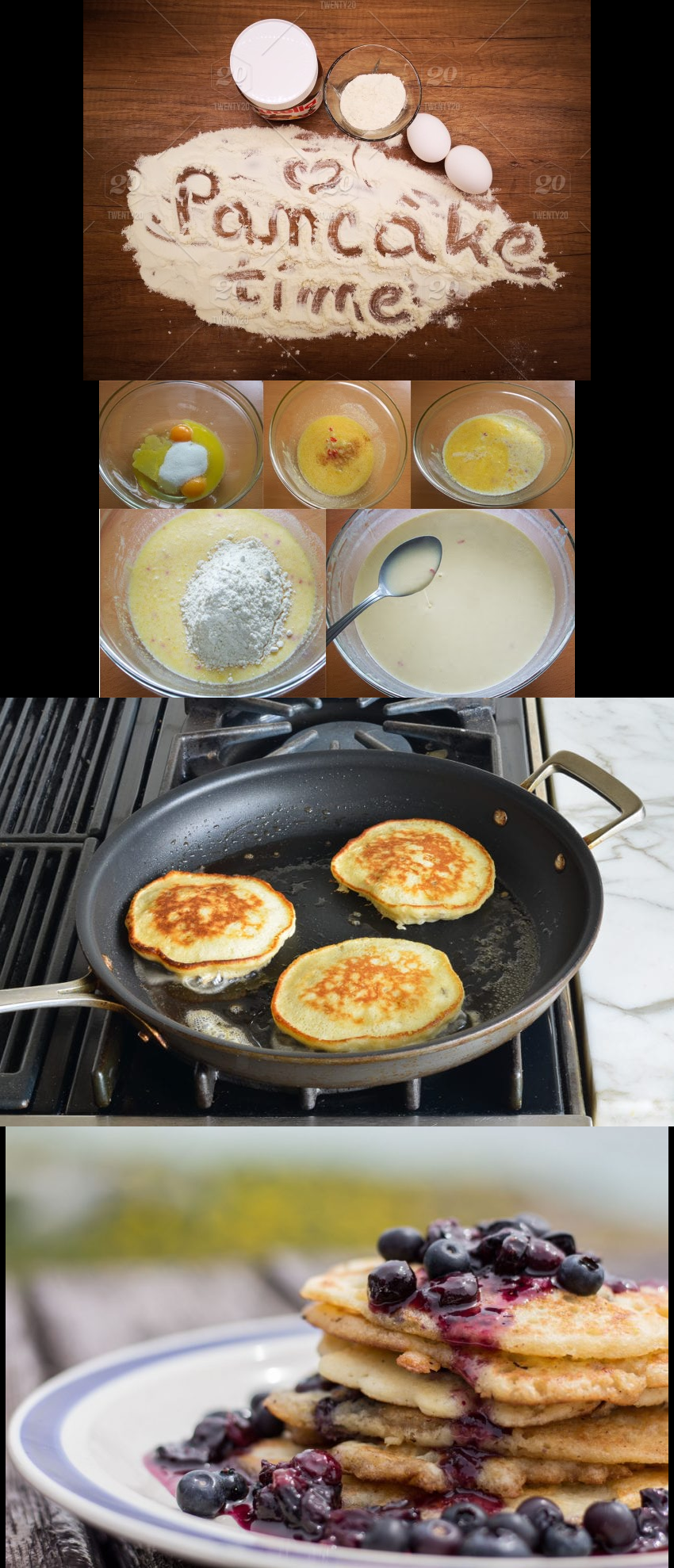}
        \caption*{(b) Right stack}
    \end{minipage}%
    \hfill
    \begin{minipage}{0.34\textwidth}
        \caption{Example of a Flip challenge from the Idena blockchain. The user is given 4 images presented in two different orderings (also referred to as stacks), and the user must select which stack of images tells a meaningful story. In this example, the answer is the right stack with the story of taking flour, mixing it with other ingredients, frying the dough, and then getting pancakes. Since this is a coherent story, then ``right'' is the correct answer.}
        \label{fig: example flip}
    \end{minipage}
    \vspace{-.2cm}
\end{figure}
Thus, there is a pressing need for benchmarks to evaluate reasoning holistically, particularly in multimodal contexts where visual and logical reasoning are required.

To address these challenges, we introduce the \textit{FLIP dataset}, derived from the Idena blockchain. Idena is a decentralized identity verification system that uses human-generated tasks, known as FLIP challenges, to verify users' ``humanness.'' In these challenges, participants are presented with two different orderings of images and must determine which ordering forms a coherent story. These tasks emphasize sequential reasoning and visual storytelling, combining the need for logical and multimodal reasoning. Humans create and verify the tasks through consensus voting, ensuring high-quality annotations. The tasks are designed to be easily solved by humans, as indicated by the fact that out of the 84,600 participant answers, 95.3\% of them are correct.

One of the key features of the FLIP dataset is its transparency and simplicity, which allow researchers to analyze model failures effectively. The task design--selecting the correct ordering--provides clear ground truth and makes it straightforward to diagnose where models may fail. For example, researchers can assess whether errors stem from misunderstanding the visual content, failing to evaluate logical coherence, or both. This contrasts with more abstract benchmarks, where identifying the root causes of failure can be more challenging, such as requiring mathematical domain knowledge. Furthermore, the FLIP dataset favors short and logical stories, requiring models to assess parameters like sequence coherence and narrative simplicity. This is important to test based on the findings by \citet{pascual2021plug} showing that LLMs can generate stories with arbitrary sets of words.

Experiments with state-of-the-art models show that reasoning through text input, such as captions generated by image captioning models, performs better than reasoning directly from visual inputs. This highlights the limitations of current AI models in processing raw visual data effectively. Meanwhile, the use of caption-based reasoning ensures that models encounter tasks with many variations rather than relying on a single instance; this should reduce the risk of models having previously encountered and memorized the exact FLIP challenges \cite{mirzadeh2024gsmsymbolicunderstandinglimitationsmathematical}.

We introduce the \textit{FLIP dataset}, a novel benchmark designed to evaluate AI reasoning capabilities in the context of visual storytelling and event-sequence comprehension. Our work comprehensively evaluates state-of-the-art models, comparing their performance on FLIP challenges using direct visual inputs versus text-based descriptions from captioning models. While FLIP’s intuitive design facilitates error analysis, our experiments reveal significant limitations in existing reasoning models and techniques, such as exemplars at inference and task reframing, when applied to FLIP challenges. Even the best-performing open-sourced models achieve maximum accuracy of 75.5\% in zero-shot settings, compared to 95.3\% achieved by human users. These findings underscore the need for robust, multimodal reasoning benchmarks and highlight the gap between current AI capabilities and human-level reasoning.

\section{Related Work}
\paragraph{Image Captioning}

Image captioning, as a fundamental task in computer vision, has seen significant advancements with the introduction of models like CLIP \cite{radford2021learningtransferablevisualmodels}, which pioneered the ability to bridge the understanding of text and images through natural language supervision. CLIP aligns visual and textual representations in a shared embedding space, enabling robust performance across various multimodal tasks. BLIP-2 \cite{blip2} extended this paradigm by integrating frozen image encoders with large language models, improving multimodal understanding. Similarly, Flamingo by \citep{alayrac2022flamingovisuallanguagemodel} demonstrated strong few-shot learning capabilities by leveraging large-scale pretraining on multimodal datasets. Earlier works, such as Conceptual Captions, laid the foundation by providing large-scale datasets for training captioning models \cite{sharma2018conceptual,suhr2019corpusreasoningnaturallanguage}, and improving how models learn from image multi-modal data \cite{cho2021unifyingvisionandlanguagetaskstext}. However, despite these advances, captioning models often struggle to generate contextually rich descriptions necessary for reasoning tasks \cite{cornia2020meshedmemorytransformerimagecaptioning}. 

\paragraph{Large Language Models and Vision-Language Models}

The integration of vision and language has propelled the development of large models capable of multimodal reasoning. \textsc{GPT-3} \citep{brown2020languagemodelsfewshotlearners} showcased remarkable language generation capabilities, which were extended to multimodal tasks with models like \textsc{PaLM-E} \citep{driess2023palmeembodiedmultimodallanguage}, an embodied language model that combines perception and action. \textsc{GPT-4}’s multimodal variant \citep{openai2024gpt4technicalreport} introduced enhanced capabilities for understanding visual inputs, demonstrating emergent reasoning abilities in vision-language tasks. \citet{bommasani2022opportunitiesrisksfoundationmodels} highlighted the potentials and risks of such foundation models, particularly in bridging vision and language modalities. Furthermore, models like \textsc{Qwen-VL} \citep{qwen2VL} have made strides in aligning vision and language components, but challenges remain in achieving seamless integration for tasks requiring deep reasoning.

\paragraph{Reasoning Models}

Models explicitly designed for reasoning tasks have focused on addressing the limitations of general-purpose LLMs and VLMs. For instance, GSM8K \citep{cobbe2021trainingverifierssolvemath}, MATH \citep{hendrycks2021measuringmathematicalproblemsolving}, and FrontierMATH \citep{glazer2024frontiermathbenchmarkevaluatingadvanced} are benchmarks introduced to evaluate mathematical and logical reasoning capabilities, highlighting the gap between human and model performance. \citet{johnson2016clevrdiagnosticdatasetcompositional} introduced CLEVR, a diagnostic dataset for compositional language and visual reasoning, emphasizing the need for models to handle fine-grained reasoning tasks. \citet{saxton2019analysingmathematicalreasoningabilities} explored the abstraction capabilities required for mathematical reasoning, while \citet{mirzadeh2024gsmsymbolicunderstandinglimitationsmathematical} demonstrated the fragility of reasoning performance in LLMs hinting at risk of exposure to data contamination. These studies underscore the need for novel datasets, like FLIP, to test reasoning abilities in a multimodal context.

\paragraph{Techniques for Improving Reasoning in Language and Vision Models}

Numerous techniques have been proposed to enhance reasoning capabilities in large models. \citet{wei2023chainofthoughtpromptingelicitsreasoning} introduced Chain-of-thought (CoT) prompting enabling models to perform step-by-step reasoning, significantly improving performance on complex tasks. \citet{nye2021workscratchpadsintermediatecomputation} introduced scratchpads to allow models to show their work during intermediate computation. \citet{kojima2023largelanguagemodelszeroshot} demonstrated that large models could act as zero-shot reasoners when appropriately prompted with exemplars. Meanwhile, \citet{zhou2023leasttomostpromptingenablescomplex} proposed least-to-most prompting to decompose tasks into manageable steps. However, these techniques often come with computational costs and scalability trade-offs, sometimes showing only minor improvements \citep{snell2024scalingllmtesttimecompute}.
\citet{khot2021textmodularnetworkslearning} explored modular networks to address these challenges, enabling efficient task decomposition for reasoning tasks. Very recently, \citet{qi2024mutualreasoningmakessmaller,guan2025rstarmathsmallllmsmaster} showed a method to improve the reasoning performance of small language models, yielding results comparable to o1 from OpenAI. 
These techniques, while effective, highlight the need to balance computational efficiency with reasoning performance.

\paragraph{Common Sense Reasoning Benchmarks}
Benchmarks for evaluating common sense reasoning, such as CommonsenseQA \citep{talmor2019commonsenseqaquestionansweringchallenge} and SocialIQA \citep{sap2019socialiqacommonsensereasoningsocial}, have driven progress in this domain. 
\citet{clark2018thinksolvedquestionanswering} introduced a challenging dataset ``AI2 Reasoning Challenge'' designed to test reasoning abilities beyond simple retrieval, focusing on scientific and commonsense reasoning. MMMU \citep{yue2024mmmumassivemultidisciplinemultimodal} extends the evaluation scope by integrating multimodal and multidisciplinary reasoning tasks. \citet{chollet2019measureintelligence} introduced in his landmark paper ``On the Measure of Intelligence'' a perspective on evaluating reasoning and intelligence in AI, offering a framework that informs the design of modern benchmarks. 
PIQA \citep{bisk2019piqareasoningphysicalcommonsense} introduced a dataset for physical commonsense reasoning, while the Winograd Schema Challenge \citep{levesque2012winograd} focused on linguistic ambiguities. HellaSwag \citep{zellers2019hellaswagmachinereallyfinish} challenged models with adversarial datasets designed to evaluate their ability to complete sentences logically. These benchmarks have, over time, taught researchers many lessons and been key to developing the field of AI \citep{kocijan2023defeatwinogradschemachallenge}. The goal of the FLIP dataset is, in part, to provide a new way to evaluate reasoning capabilities in a multimodal setting incorporating sequential logic and visual context.

\section{Methodology}

\subsection{Data}
\paragraph{What flip challenges are}
The flip challenges from Idena consist of 4 images presented in two orderings, left and right; see \Cref{fig: example flip} for an example. The user is then asked to select the ordering that tells a meaningful story \cite{idena_flip_challenge}. The flip challenges were solved by human users, who could choose either left, right, or to report the flip. The users are asked to report flips that do not follow the design guidelines, for instance, if the images contain numbers indicating the ordering. 

So, for each flip challenge, we also have how the users voted on it, i.e., how many voted left, how many voted right, and how many reported the flip. Based on the votes, the flips are given a consensus score of either ``No consensus,'' ``Weak consensus,'' or ``Strong consensus.'' This scoring is done by the blockchain protocol and is given along with the votes.

\paragraph{Collected data}
The data was acquired by parsing the available portion of the challenges on Idena's website~\cite{idena_explorer}. At the time of writing, the Idena blockchain consists of 153 epochs, each consisting of between 100 and 50,000 flips. However, no flips are available from the first 5 epochs, and all flips after the 36th epoch are encrypted, meaning that some of the images are hidden. Thus, we can collect data from around 30 epochs where we discard any flips that have ``No consensus.'' This results in 11,674 flips split into disjoint train, validation, and test sets. Since some experiments can be computationally heavy, we also take subsamples of the train, validation, and test sets, which we denote as train (short), validation (short), and test (short). The number of flips in each set can be seen in \Cref{tab: number of collected flips}.

\paragraph{Dataset statistics}
The statistics of the collected dataset are summarized in \Cref{tab: data statistics}. The table shows that most flip challenges have strong consensus (95.7\%), with only a minor fraction categorized as weak consensus (4.3\%). The user solutions are nearly evenly distributed between the Left (49.4\%) and Right (50.6\%) orderings, indicating no significant bias toward either option. This balance and the high proportion of strong consensus flips suggest that users generally agree on the meaningful story conveyed by the images in the flip challenges.

\subsection{Requirements of a Successful Model}
The Idena blockchain defines ``Verified'' users as users correctly solving flip challenges at least 75\% of the time and ``Human'' users as users solving at least 92\% of the flip challenges correctly and having solved at least 24 flip challenges \cite{idena_whitepaper}.\footnote{When looking over our entire dataset we get that users on average solve 95.3\% the challenges correctly.}
Furthermore, the Idena team incentivizes an open-sourced AI tool that can solve flip challenges with an accuracy between 71\% and 80\% (rounded down to the nearest integer).
Based on this, we define the minimum performance criteria for a ``successful model'' as accuracy above 71\% with human-level performance being above 95\%.

Since the reward requires entirely open-sourced tools, we focus primarily on open-source generative models. We will also test closed-sourced models such as OpenAI's GPT4o and Google's Gemini 1.5 Pro, but fewer experiments will be done with these. This choice is motivated by the requirements for qualifying for the reward and by promoting the use and research of open-sourced models.

\subsection{Models}
We first tried some baseline models using ResNet50 \cite{he2015deepresiduallearningimage} to get embeddings of the images and then used MLPs and k-NNs for predictions.
However, these models only achieve an accuracy of around 60\%, which is far below the minimum required performance of a successful model of 71\% as random guessing would get $\approx$50\%. See \Cref{appendix: mlps and knns} for full results.

\paragraph{Pipeline}
We then employ various state-of-the-art generative language models. For them to handle the input images, we must either use vision-language models (VLMs) or a captioning model such as BLIP before feeding the generated image captions to a (large) language model (LLM). In the latter setup, we call the LLM the reasoning model.

\begin{table}[t]
    \begin{minipage}{0.49\textwidth}
        \centering
        \scriptsize
        \begin{tabular}{c|c}
            \toprule
            Data set & Number of flips \\
            \midrule
            Train (short) & 490 (4.2\%)\\
            Validation (short) & 105 (0.9\%)\\
            Test (short) & 106 (0.9\%)\\
            \midrule
            Train & 3502 (30\%)\\
            Validation & 3502 (30\%)\\
            Test & 4670 (40\%)\\
            \midrule
            Total flips (full) & 11674 \\
            \bottomrule
        \end{tabular}
        \caption{The number of collected flip challenges from the Idena blockchain. The dataset consists of 11,674 flips, divided into train, validation, and test sets. We create small subsets, marked with ``(short)'', that allow experimentation with computationally intensive models.}
        \label{tab: number of collected flips}
    \end{minipage}
    \hfill
    \begin{minipage}{0.49\textwidth}
        \centering
        \scriptsize
        \renewcommand{\arraystretch}{1.6} 
        \setlength{\tabcolsep}{3.5pt}
        \begin{tabular}{cc|cc|c}
            \toprule
            & & \multicolumn{2}{c|}{Consensus} &  \\
            & & Weak & Strong & Totals\\
            \midrule
            \parbox[t]{3.7mm}{\multirow{2}{30pt}{\rotatebox[origin=c]{90}{Solution}}} & Left & 250 (2.1\%) & 5516 (47.2\%) & 5766 (49.4\%) \\
            & Right & 250 (2.1\%) & 5658 (48.4\%) & 5908 (50.6\%) \\
            \midrule
            & Totals & 500 (4.3\%) & 11174 (95.7\%) & 11674 (100.0\%) \\
            \bottomrule
        \end{tabular}
        \caption{The distribution of user solutions for flip challenges in the dataset. Each flip challenge has a consensus level (Weak/Strong) and correct ordering (Left/Right). The table includes the totals for each ordering and consensus level and overall percentages relative to the total dataset.}
        \label{tab: data statistics}
    \end{minipage}
\end{table}
Image descriptions were generated by captioning models and excluded from the input tokens. All captioning models received equivalent prompts for each image, with a minor exception for BLIP2 models where the standard prompt was modified to avoid generic responses like ``Yes, I can.'' 
The prompts given to the captioning and reasoning models are provided in \Cref{tab: prompts}. We will also do an experiment where we change how we prompt the reasoning models to see the effect of this. This updated prompt is also given in the table. 

The experimental pipeline remained consistent across all models and most experiments, with captioning models describing image content and reasoning models receiving standardized task interpretations, captions, and answer formatting instructions.
In the standard reasoning prompt, after the reasoning prompt from \Cref{tab: prompts}, the two stories are presented as ``Story 1: $\backslash$n Image 1 caption $\backslash$n Image 2 caption ... Image 4 caption $\backslash$n$\backslash$n Story 2: Image 1 caption ...''
\footnote{We provide the full codebase and data when publishing the paper.}

The following subsections list the captioning, reasoning (open- and closed-sourced), and VL models we considered. 

\begin{table}[t]
    \centering
    \tiny
    \setlength{\tabcolsep}{3.0pt}
    \begin{tabular}{p{0.18\columnwidth}|p{0.75\columnwidth}}
        \toprule
        \multicolumn{2}{c}{Captioning} \\
        \midrule
        General prompt & Can you please describe this image? \\
        \midrule
        BLIP prompt & What do you see in this image? Please describe it \\
        \midrule
        \multicolumn{2}{c}{Reasoning} \\
        \midrule
        General prompt & You are given 2 Stories comprised of 4 images each. The images are described by captions generated by some random image captioning model and may not reflect very accurate image descriptions, but this is all we have. One of the Stories is logically ordered and capable of conveying a simple story, whereas the other is not. Please determine which story (1 or 2) is more logically ordered and provide the result as either '1' or '2' after 'Solution:' \\
        \midrule
        Modified prompt & You are given 4 images: The 4 captions are generated by some random image captioning model and may not reflect very accurate image description, but this is all we have. And there are 2 possible options to order these images: One of the options presents a logical and coherent sequence and is capable of conveying a simple story, whereas the other is less so. Please determine which order (1 or 2) is more likely the correct one and provide the result as either '1' or '2' \\
        \bottomrule
    \end{tabular}
    \caption{Prompts given to the captioning and reasoning models when solving FLIP challenges. We have a general captioning prompt that worked for most models, but the BLIP models would sometimes output ``Yes, I can.,'' so we modified the prompt for them. We also have a standard reasoning prompt used in most experiments and a modified variant used in an experiment to determine the effect of prompting differently.}
    \label{tab: prompts}
    \vspace{-.3cm}
\end{table}

\subsubsection{Captioning models}
We considered 9 captioning models across 4 architectures, namely \textsc{VipLlava} (\textsc{7B} \& \textsc{13B})~\citep{vipllava}, \textsc{LlavaNeXT} (\textsc{Mistral 7B}, \textsc{Vicuna 7B}, \textsc{Vicuna 13B})~\citep{llavanext}, \textsc{BLIP2} (\textsc{2.7B}, \textsc{6.7B}, \textsc{Flan T5 XXL})~\citep{blip2}, and \textsc{Llama 3.2} (Vision - Instruct \textsc{11B}) ~\citep{grattafiori2024llama3herdmodels}. Note that \textsc{BLIP2 Flan T5 XXL} has 12B parameters. 

\subsubsection{Open sourced reasoning models}
For the open-sourced reasoning models we run Llama 3.1 70B from Meta (\textsc{Meta Llama 3.1 70B}) ~\citep{grattafiori2024llama3herdmodels}, \textsc{Qwen 2.5} ~\citep{qwen2025qwen25technicalreport}, \textsc{Qwen 2 VL} ~\citep{qwen2VL} (both loaded in 8bit) and Llama 3.1 Nemotron 70B \& 51B from Nvidia (\textsc{NVM Llama3 51B} and \textsc{NVM Llama3 70B}) ~\citep{nvm_llama}.

We experimented with \textsc{Grin MoE} ~\citep{grin_moe} and \textsc{Phi 3.5 MoE} ~\citep{abdin2024phi3technicalreporthighly}, but they both gave poor results (accuracies around 50\%). Thus, we omit them in the main results but include results for them in the appendix.

\subsubsection{Closed sourced reasoning models}
We also tested closed-sourced models like GPTs from OpenAI and Geminis from Google. However, as these are much more expensive to run than the open-sourced models, we first tested a wide selection of models on the short train and validation sets. This result can be seen in \Cref{tab: closed-sourced short}. The table shows that the GPT 4 turbo model from OpenAI is the best closed-sourced model tested and performs at least 1\%-point better than all other closed-sourced models. The Gemini Flash model is relatively cheap to run on the experiments, so we include both Gemini models in further experiments. 

Based on these results, we will run future experiments on \textsc{Gemini 1.5 Flash} 002 and \textsc{Gemini 1.5 Pro} 002 ~\citep{team2023gemini}, and \textsc{GPT 4 turbo} 2024-04-09 ~\citep{openai2023chatgpt}. The versions were the newest available at the time of writing. 

\begin{table}[t]
    \begin{minipage}{0.49\textwidth}
        \centering
        \footnotesize
        \begin{tabular}{c|c}
            \toprule
            Model & Accuracy \\
            \midrule
            \multicolumn{2}{c}{OpenAI's GPT models} \\
            \midrule
            GPT 4o & 74.8\% \\
            GPT 4 turbo & \textbf{76.8\%} \\
            GPT 4 & 75.1\% \\
            GPT 4 mini & 67.7\% \\
            GPT 3.5 turbo & 54.28\% \\
            \midrule
            \multicolumn{2}{c}{Google's Gemini models} \\
            \midrule
            Gemini 1.5 Pro & \textbf{64.0\%} \\
            Gemini 1.5 Flash & 60.1\% \\
            \bottomrule
        \end{tabular}
        \caption{Accuracy of closed-sourced models when evaluated on the short train and validation sets (595 flips) using \textsc{ViPLlava 7B} for captions. We have highlighted the highest accuracies in bold for OpenAI's and Google's models.}
        \label{tab: closed-sourced short}
    \end{minipage}
    \hfill
    \begin{minipage}{0.47\textwidth}
        \centering
        \footnotesize
        \begin{tabular}{c|c}
            \toprule
            Single stacked image per story & 69.6\% \\
            4 images per story & 64.6\% \\
            \bottomrule
        \end{tabular}
        \caption{Evaluation of \textsc{Gemini 1.5 Pro} with images as the input instead of captions. In the ``single stacked image per story,'' the four images are stacked into a single image, while in the other, the 4 images are given independently. The former performs better but is still worse than using captions.}
        \label{tab: Gemini vlm}
    \end{minipage}
\end{table}

\paragraph{Why no \textsc{GPT o1}?} 
While we aimed to include as many relevant models as possible, \textsc{GPT o1} was excluded from our evaluation due to persistent technical issues and prohibitive computational costs. Nonetheless, our inclusion of a wide range of cutting-edge models supports the robustness and generalizability of our findings.

\subsubsection{Vision-Language Models}
We also test that some of our reasoning models can take images as input instead of text. The results for this evaluation are seen in \Cref{tab: Gemini vlm}, and we overall see decent results when comparing with the numbers in \Cref{tab: closed-sourced short}. However, as we will see later, the results are still worse than when using captions from BLIP. Therefore, we do not run exhaustive experiments on VLMs but leave this area for future work.

\subsubsection{Risk of Data Contamination}
As mentioned earlier, \citet{mirzadeh2024gsmsymbolicunderstandinglimitationsmathematical} show how the reasoning performance of many commonly used models drops on a standard reasoning benchmark like \textsc{GSM8K}, which they argue is due to data contamination. Therefore, asking if our setup also has this risk is prudent. The data is freely available on the internet, so there is a possibility the models have seen it before. 
For instance, when asking ChatGPT: ``Give a short and condensed answer to this question. What is the Idena blockchain, and what are the flip challenges it uses?'' It then answers: ``Idena is a blockchain that uses proof-of-personhood, requiring participants to verify their uniqueness and humanity through "flip challenges." Flips are CAPTCHAs created by users, consisting of story-based image sequences, where others must identify the correct logical order to validate the challenge. This ensures decentralized and Sybil-resistant identity verification.''

Thus, ChatGPT knows of FLIP challenges. However, this should not be a problem as we use the captioning models. The captions restate the challenges in several ways and should give a similar effect to the modifications by \citet{mirzadeh2024gsmsymbolicunderstandinglimitationsmathematical}. The caption changes should be effective since most tested models do not have a vision component. Furthermore, we make no references to Idena or FLIP challenges in the prompts, making the models less likely to connect a problem to the specific flip challenge. 
Based on this, we argue data contamination is unlikely to affect the results.


\subsection{Experiments}\label{sec: experiments}
We conducted four experiments to evaluate language models' reasoning capabilities on the FLIP challenge. In the first experiment, we ran all the models we considered, evaluating them broadly on a general query. Subsequent experiments focus on models that did well in prior experiments. Due to space constraints, we move some details to \Cref{appendix: methodology}.


\subsubsection{General Benchmark}
The first experiment assessed the zero-shot performance of large language models (LLMs) on the dataset. It focused on evaluating a large set of reasoning and captioning models to get a broad impression of the results. 


\subsubsection{Summarization of Captions}
Some captioning models provide much more extended captions than others, as shown in \Cref{tab: caption examples,fig: caption lengths distributions}. This gives extra details but might also impact the models negatively as there might be a lot of redundant or misleading information, so the vital information is lost. Hence, we designed an experiment to try to circumvent this issue by asking a language model to summarize the captions and then running the evaluation using these summarized captions. 
Specifically, we first prompt the captioning models with the extended prompt seen in \Cref{tab: summary prompts}, and we then ask a language model (the reasoning models such as \textsc{Meta Llama 3.1 70B}) to summarize the output from the captioning models. \Cref{tab: caption examples} also includes examples of the summarized extended captions for \Cref{fig: example image for captions}.

\subsubsection{Task Reframing}
The third experiment shifted focus from story cohesion to sequence meaningfulness. Images were assigned letters (A, B, C, D) and presented with their descriptions, for example: "image A: sliced apple on a cutting board." The models were then presented with two possible ordering options (e.g., Order 1: B, C, D, A and Order 2: A, C, B, D) and asked to select the more logically coherent sequence.

\subsubsection{Inference with Historical Context}
Our fourth experiment drew inspiration from the approach by \citet{wei2022finetunedlanguagemodelszeroshot} of giving examples at inference. The idea is to see if the models benefit from seeing how they previously answered FLIP challenges correctly and incorrectly. 

For this purpose, we generated contextual data using two reasoning models, \textsc{Qwen 2.5} and \textsc{Gemini 1.5 Pro}, with captions from \textsc{BLIP2 Flan T5 XXL}. We run these two combinations on 1k flip challenges from the train set using the setup in the general benchmark. The results were then marked as correct or incorrect. 
Using these samples, we constructed balanced contexts of varying lengths for inference on the validation data. When using a context length of, e.g., 8, the reasoning model is given 4 FLIP challenges with solutions and told it solved them correctly, and 4 FLIP challenges with solutions and told it solved them incorrectly. 

\begin{table*}[t]
    \centering
    \scriptsize
    \setlength{\tabcolsep}{1.4pt}
    \begin{tabular}{l|rr|rrr|rrr|r}
        \toprule
        & \multicolumn{2}{c|}{\textsc{ViPLlava}} & \multicolumn{3}{c|}{\textsc{LlavaNeXT}} & \multicolumn{3}{c|}{\textsc{BLIP2}} & \\
         & \textsc{7B} & \textsc{12B} & \textsc{Mistral 7B} & \textsc{Vicuna 7B} & \textsc{Vicuna 13B} & \textsc{2.7B} & \textsc{6.7B COCO} & \textsc{Flan T5 XXL} & \textsc{Llama 3.2 11B} \\
        \midrule
        \textsc{Qwen 2 VL} & 57.2\% & 56.7\% & 54.3\% & 55.0\% & 54.1\% & 64.1\% & 65.5\% & 67.6\% & 59.6\% \\
        \textsc{Qwen 2.5} & 66.6\% & 66.0\% & \textbf{62.9\%} & \textbf{57.7\%} & \textbf{58.3\%} & \textbf{69.8\%} & 70.5\% & 72.8\% & \textbf{61.5\%} \\
        \textsc{NVM Llama3 51B} & 57.0\% & 57.2\% & 50.3\% & 50.4\% & 49.6\% & 69.4\% & 71.0\% & \textbf{73.7\%} & 48.5\% \\
        \textsc{NVM Llama3 70B} & \textbf{67.1\%} & \textbf{67.2\%} & 59.6\% & 56.4\% & 55.2\% & 69.1\% & 70.4\% & 72.5\% & 58.5\% \\
        \textsc{Meta Llama 3.1 70B} & 65.5\% & 65.7\% & 57.6\% & 53.6\% & 52.7\% & 69.7\% & \textbf{71.2\%} & 73.1\% & 58.8\% \\
        \midrule
        \textsc{Gemini 1.5 Flash} & 61.5\% & 61.7\% & 53.3\% & 52.7\% & 50.0\% & 69.2\% & 67.1\% & 69.8\% & 51.8\% \\
        \textsc{Gemini 1.5 Pro} & \textbf{68.9\%} & \textbf{67.8\%} & \textbf{67.6\%} & \textbf{66.2\%} & \textbf{65.2\%} & \textbf{71.7\%} & \textbf{71.1\%} & 75.2\% & \textbf{68.0\%} \\
        \textsc{GPT 4 Turbo} & NA & NA & NA & NA & NA & NA & NA & \textbf{77.9\%} & NA \\
        \bottomrule
    \end{tabular}
    \caption{Accuracies for the tested models in the general experiment. We test a broad range of captioning and reasoning models; the row indicates the reasoning model used, and the columns indicate the captioning model used. We only tested \textsc{GPT 4 Turbo} on the captions from \textsc{BLIP2 Flan T5 XXL} as it was the best-performing captioning model for all other tested reasoning models. In bold, we highlight each captioning model's best open-sourced and closed-sourced reasoning models. We include \Cref{tab: general results full} in the appendix with the results for all tested reasoning models.}
    \label{tab: general results partial}
\end{table*}

\begin{table*}[t]
    \centering
    \tiny
    \setlength{\tabcolsep}{3.5pt}
    \begin{tabular}{l|ccc|ccc|ccc}
        \toprule
        Summarization model & \multicolumn{3}{c|}{\textsc{NVM Llama3 51B}} & \multicolumn{3}{c|}{\textsc{Meta Llama 3.1 70B}} & \multicolumn{3}{c}{\textsc{Qwen 2.5}} \\
        \midrule
        \multirow{2}{*}{Captioning model} & \multicolumn{2}{c}{\textsc{ViPLlava}} & \multicolumn{1}{c|}{\textsc{BLIP2}} & \multicolumn{2}{c}{\textsc{ViPLlava}} & \multicolumn{1}{c|}{\textsc{BLIP2}} & \multicolumn{2}{c}{\textsc{ViPLlava}} & \multicolumn{1}{c}{\textsc{BLIP2}}\\
         & \textsc{7B} & \textsc{12B} & \textsc{Flan T5 XXL} & \textsc{7B} & \textsc{12B} & \textsc{Flan T5 XXL} & \textsc{7B} & \textsc{12B} & \textsc{Flan T5 XXL} \\
        \midrule
        \textsc{Qwen 2.5} & 70.5 (4.1) & 72.9 (6.6) & 73.3 (-0.1) & 69.7 (3.2) & 70.6 (4.3) & 70.9 (-2.6) & 71.4 (4.9) & 72.5 (6.2) & 72.3 (-1.1) \\
        \textsc{NVM Llama3 51B} & 68.5 (11.6) & 68.9 (11.7) & 70.2 (-3.5) & 66.7 (9.8) & 65.3 (8.1) & 70.9 (-2.8) & 68.2 (11.3) & 69.2 (12.0) & 71.3 (-2.5) \\
        \textsc{NVM Llama3 70B} & 70.7 (3.2) & 72.1 (4.5) & 70.9 (-1.8) & NA & NA & 71.4 (-1.3) & 71.2 (3.7) & 71.8 (4.2) & 70.2 (-2.5) \\
        \textsc{Meta Llama 3.1 70B} & 72.1 (5.5) & 71.7 (4.9) & 72.9 (-0.2) & 69.3 (2.7) & 71.3 (4.5) & 72.8 (-0.3) & 71.9 (5.2) & 72.7 (5.9) & 72.2 (-0.8) \\
        \bottomrule
    \end{tabular}
    \caption{Accuracies (in \%) after summarizing the captions provided by various captioning models. We indicate in parenthesis the gain the models see in accuracy over the numbers in \Cref{tab: general results partial}. There are significant improvements for models using summarized captions from \textsc{ViPLlava}; however, the captions from \textsc{BLIP2 Flan T5 XXL} are not better.}
    \label{tab: summarization}
\end{table*}

\section{Results}
This section presents the results of the experiments mentioned in \Cref{sec: experiments}. Finally, we will analyze the error of the best-performing models. Due to limited space, we move the error analysis and detailed results for an ensemble model into the appendix, cf. \Cref{appendix: extended results}. Most relevantly, we see that a simple ensemble of 5 open-sourced models can get an accuracy of 80.1\%.

\subsection{General Benchmark}
We see the results of the general benchmark in \Cref{tab: general results partial} with a few reasoning models omitted; see \Cref{tab: general results full} for all models. None of the reasoning models can get accuracies above the minimum performance criteria of 71\% when using the \textsc{ViPLlava}, \textsc{LlavaNeXT}, or \textsc{Llama} captioning models. 
\begin{table*}[t]
    \centering
    \tiny
    \setlength{\tabcolsep}{3.0pt}
    \begin{tabular}{l|rr|rrr|rrr|r}
        \toprule
        & \multicolumn{2}{c|}{\textsc{ViPLlava}} & \multicolumn{3}{c|}{\textsc{LlavaNeXT}} & \multicolumn{3}{c|}{\textsc{BLIP2}} & \\
         & \textsc{7B} & \textsc{12B} & \textsc{Mistral 7B} & \textsc{Vicuna 7B} & \textsc{Vicuna 13B} & \textsc{2.7B} & \textsc{6.7B COCO} & \textsc{Flan T5 XXL} & Means\\
        \midrule
        \textsc{Qwen 2.5} & 66.6 (0.2) & 66.5 (0.2) & 63.6 (0.1) & 55.0 (-4.5) & 58.0 (-1.3) & 70.8 (0.5) & 71.3 (0.8) & 73.7 (0.2) & 65.7 (-0.5)\\
        \textsc{NVM Llama3 51B} & 67.2 (10.3) & 66.8 (9.6) & 56.4 (6.7) & 52.6 (2.7) & 55.6 (6.8) & 71.3 (1.4) & 73.4 (2.4) & 75.5 (1.7) & 64.8 (5.2)\\
        \textsc{NVM Llama3 70B} & 67.7 (0.2) & 67.2 (-0.4) & 68.8 (9.3) & 60.9 (3.6) & 58.9 (3.8) & 66.5 (-3.1) & 68.4 (-2.1) & 70.9 (-1.8) & 66.2 (1.2)\\
        \textsc{Meta Llama 3.1 70B} & 68.1 (1.4) & 68.4 (1.6) & 69.8 (11.4) & 58.9 (3.3) & 58.7 (5.4) & 66.8 (-3.3) & 68.2 (-3.1) & 71.5 (-1.6) & 66.3 (1.9)\\
        \midrule
        Means & 67.4 (3.0)  & 67.2 (2.7)  & 64.6 (6.9)  & 56.9 (1.3)  & 57.8 (3.7)  & 68.9 (-1.1)  & 70.3 (-0.5)  & 72.9 (-0.3) & 65.8 (2.0) \\
        \bottomrule
    \end{tabular}
    \vspace{-.15cm}
    \caption{Accuracies (in \%) for 4 reasoning and 8 captioning models when reframing the task along with means across columns, rows, and all values. We indicate in parenthesis the gain the models see in accuracy over the numbers in \Cref{tab: general results partial}. Notably, we see some high increase of 9 to 10\%-points for several model pairs, but these combinations did not do well in the general benchmark. On average, the gain is 2.0\%-points, showing that the reframed task is better.}
    \label{tab: task reframing}
    \vspace{-.15cm}
\end{table*}

\subsection{Summarization of Captions}
\Cref{tab: summarization} has the results of summarizing the captions provided by 3 captioning models. The results show that the accuracies when using captions from \textsc{ViPLlava} are much better, improving by 12\%-points in one instance. 
However, the captions from \textsc{BLIP2 Flan T5 XXL} are not better. The BLIP captions are generally very short, so they might not have the information in the first place, making the summarization irrelevant. Overall, the results when using summarized models are only around the minimum performance criteria, and several better models do worse.

\begin{table}[t]
    \centering
    \footnotesize
    \setlength{\tabcolsep}{4.0pt}
    \begin{tabular}{l|ccccc}
         \toprule
         & \multicolumn{5}{c}{Context window size} \\
         & 0 & 2 & 4 & 8 & 16 \\
         \midrule
         \textsc{Qwen 2.5} & 72.4\% & 68.6\% & 70.2\% & 67.7\% & 70.9\% \\
         \midrule
         \textsc{Gemini 1.5 Pro} & 75.2\% & 72.1\% & 66.7\% & 68.3\% & 69.5\% \\
         \bottomrule
    \end{tabular}
    \vspace{-.15cm}
    \caption{Accuracies for 2 of the best models with context of previous answers on the training data. The models use captions from \textsc{BLIP2 Flan T5 XXL}. The context window of size 0 corresponds exactly to the setting in the general benchmark, so the numbers in that column are from \Cref{tab: general results partial}.}
    \label{tab: context}
    \vspace{-.15cm}
\end{table}
\subsection{Task Reframing}
We see the results of the task reframing in \Cref{tab: task reframing}. Overall, reframing the task is an advantage with a mean gain of 2.0\%-points over the general benchmark results in \Cref{tab: general results partial} for the tested models.

\subsection{Inference with Historical Context}
\Cref{tab: context} shows the results of giving \textsc{Qwen 2.5} and \textsc{Gemini 1.5 Pro} historical context (exemplars). We tested sizes from 2 to 16 and included results from the general benchmark for reference, as the general benchmark corresponds to a context size of 0. Both models do worse when given the context of previous answers. Especially the performance of \textsc{Gemini 1.5 Pro} drops when given $>2$ exemplars.

\section{Conclusion}
This work presents the FLIP dataset as a benchmark for evaluating reasoning capabilities in AI systems. FLIP challenges uniquely combine sequential reasoning, visual storytelling, and common sense, addressing limitations of traditional benchmarks such as narrow task domains and risks of data contamination. Human-generated and consensus-validated tasks ensure transparency and quality, making FLIP practical and interpretable.

Our experimental results demonstrate the current gap between AI and human reasoning capabilities. While state-of-the-art models benefit from textual captions to improve performance, the struggles with visual inputs underscore the limitations of existing multimodal reasoning systems. Though significant challenges remain, task reframing and ensembles promise to enhance the models.

This reasonably high reasoning performance is noteworthy, as the models do not directly process raw images, relying instead on captions generated by various image captioning models. The consistency across captioning methods suggests a low risk that the reasoning models have previously encountered and memorized these specific problems.

We also note that the Idena Blockchain's use of FLIP challenges as a security measure might be flawed, as we saw ensembles achieve $>85\%$ accuracy.

By introducing FLIP, we aim to drive progress in multimodal reasoning and provide a tool for researchers to benchmark models on tasks that mimic real-world human reasoning. Future work could refine AI systems for better generalization and explore how multimodal learning strategies can bridge the gap between AI and human-level reasoning.

\bibliography{references}
\bibliographystyle{iclr2025_conference}

\appendix
\newpage

\section{Appendix -- MLPs and k-NNs Baseline Performance}\label{appendix: mlps and knns}

Since the problem can be seen as a classification problem, and \citet{plesner2024breaking} demonstrated that classic models like YOLOv8 can solve CAPTCHAs, we first tested a similar pipeline using ResNet50 as the backbone \cite{yolov8_ultralytics,he2015deepresiduallearningimage}. Given a flip challenge, we use ResNet50 to get embeddings for each image, and we then stack the embeddings in the two orderings given by the left and right stacks and concatenate the result. This is fed to a multi-layer perceptron, MLP, or used with k-nearest neighbors to classify if the left or right stack is correct. The results can be seen in \Cref{tab: mlp and knn performance}. We see that the performance is only in the high 50s to 60s for the two methods. This is below the minimum performance of a successful model of 70\% and the performance of humans of 90\%.

\begin{table}[H]
    \centering
    \begin{tabular}{c|c}
        \toprule
        MLP & 62.3\% \\
        k-NN (k=18) & 58.9\% \\
        \bottomrule
    \end{tabular}
    \caption{Performance of the best MLP and k-NN baseline models on the flip challenges. The best k-NN uses k=18 neighbors, and the best MLP has two hidden layers of size 2000 and 100, respectively. We see that the performance is only in the high 50s to 60s for the two methods. This is below the minimum performance of a successful model of 70\% and the performance of humans of 90\%.}
    \label{tab: mlp and knn performance}
\end{table}

\section{Appendix -- Methodology}\label{appendix: methodology}

\begin{table}[H]
    \centering
    \scriptsize
    \setlength{\tabcolsep}{3.5pt}
    \begin{tabular}{p{0.28\columnwidth}|p{0.64\columnwidth}}
        \toprule
        Captioning prompt & What is the primary subject of the image? What is the relationship between subjects if there are multiple people or objects? What is the potential story or narrative implied by the scene? Are there symbols, motifs, or repeated patterns that add meaning to the image? What actions are the subjects performing, and how does that convey their purpose? \\
        \midrule
        Summarization prompt & You are given QA pairs about an image, and they appear to contain too much text to be used as captions for the image. Your task is to summarize these QA pairs in at most one sentence, retaining details that allow to recognize the exact image this caption belongs to among many similar ones.  \\
        \bottomrule
    \end{tabular}
    \caption{Extended captioning prompt given to the captioning models to get more extensive information. An LLM (here, one of the reasoning models such as \textsc{Meta Llama 3.1 70B}) is then asked to summarize the captions using the summarization prompt.}
    \vspace{-0.5cm}
    \label{tab: summary prompts}
\end{table}

\begin{table}[H]
    \centering
    \scriptsize
    \begin{tabular}{c|p{10.6cm}}
        \toprule
        \multicolumn{2}{c}{Baseline Captions} \\
        \midrule
        \textsc{ViPLlava 13B} & The image features a wooden gavel, which is a tool commonly used in courtrooms to signal the start or end of a session. The gavel is positioned on a wooden surface, likely a table, and is the main focus of the scene. In the background, there are two people partially visible. One person is on the left side of the image, while the other is on the right side. They appear to be seated and are not the main focus of the image. \vspace{4pt} \\
        \textsc{BLIP2 Flan T5 XXL} & A judge's gavel on a wooden table. \\
        \midrule
        \multicolumn{2}{c}{Summarized Extended Captions} \\
        \midrule
        \textsc{ViPLlava 13B} & A gavel rests on a wooden surface, likely a judge's desk, with two people partially visible in the background, suggesting a courtroom or legal setting. \vspace{4pt} \\
        \textsc{BLIP2 Flan T5 XXL} & A judge's gavel lies on a wooden table, symbolizing justice, as if a judge has just struck a book while deciding a case.
    \end{tabular}
    \caption{Examples of captions provided by \textsc{ViPLlava 13B} and \textsc{BLIP2 Flan T5 XXL} using the general captioning prompts in \Cref{tab: prompts} and the new summarization of extended captions for the image in \Cref{fig: example image for captions}. We see that the baseline prompts vary greatly in the length and details given, while the summarized captions are more similar.}
    \label{tab: caption examples}
\end{table}

\begin{figure}[t]
    \begin{minipage}{0.35\textwidth}
        \centering
        \includegraphics[width=0.95\linewidth]{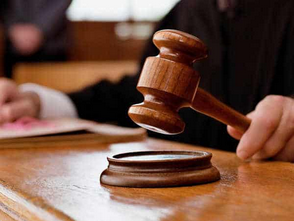}
        \caption{Image the captions in \Cref{tab: caption examples} are made for.}
        \label{fig: example image for captions}
    \end{minipage}
    \hfill
    \begin{minipage}{0.59\textwidth}
        \centering
        \includegraphics[width=0.85\linewidth]{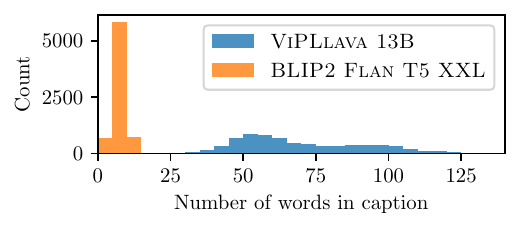}
        \caption{Number of words in the captions provided by \textsc{ViPLlava 13B} and \textsc{BLIP2 Flan T5 XXL}. We see that the former produces much longer captions than the latter.}
        \label{fig: caption lengths distributions}
    \end{minipage}
\end{figure}

\newpage
\section{Appendix -- Extended Results}\label{appendix: extended results}

\subsection{General Benchmark}
\Cref{tab: general results full} shows the results for all tested models in the general benchmark. Due to the page limit, we omitted a few models in \Cref{tab: general results partial}. The additional models are generally not good.

\begin{table*}[t]
    \centering
    \tiny
    \setlength{\tabcolsep}{3.5pt}
    \begin{tabular}{l|rr|rrr|rrr|r}
        \toprule
        & \multicolumn{2}{c|}{\textsc{ViPLlava}} & \multicolumn{3}{c|}{\textsc{LlavaNeXT}} & \multicolumn{3}{c|}{\textsc{BLIP2}} & \\
         & \textsc{7B} & \textsc{12B} & \textsc{Mistral 7B} & \textsc{Vicuna 7B} & \textsc{Vicuna 13B} & \textsc{2.7B} & \textsc{6.7B COCO} & \textsc{Flan T5 XXL} & \textsc{Llama 3.2 11B} \\
        \midrule
        \textsc{Qwen 2 VL} & 57.2\% & 56.7\% & 54.3\% & 55.0\% & 54.1\% & 64.1\% & 65.5\% & 67.6\% & 59.6\% \\
        \textsc{Qwen 2.5} & 66.6\% & 66.0\% & \textbf{62.9\%} & \textbf{57.7\%} & \textbf{58.3\%} & \textbf{69.8\%} & 70.5\% & 72.8\% & \textbf{61.5\%} \\
        \textsc{NVM Llama3 51B} & 57.0\% & 57.2\% & 50.3\% & 50.4\% & 49.6\% & 69.4\% & 71.0\% & \textbf{73.7\%} & 48.5\% \\
        \textsc{NVM Llama3 70B} & \textbf{67.1\%} & \textbf{67.2\%} & 59.6\% & 56.4\% & 55.2\% & 69.1\% & 70.4\% & 72.5\% & 58.5\% \\
        \textsc{Meta Llama 3.1 70B} & 65.5\% & 65.7\% & 57.6\% & 53.6\% & 52.7\% & 69.7\% & \textbf{71.2\%} & 73.1\% & 58.8\% \\
        \textsc{MS GRIN MoE} & 49.6\%  & 47.9\%  & 49.9\%  & 47.5\%  & 47.2\%  & 57.6\%  & 59.6\%  & 61.7\%  & 46.7\% \\
        \textsc{Phi 3.5 MoE} & 47.3\%  & 48.8\%  & 48.0\%  & 47.1\%  & 46.5\%  & 62.2\%  & 62.3\%  & 64.3\%  & 47.1\% \\
        \midrule
        \textsc{Gemini 1.5 Flash 001} & 59.4\%  & 59.2\%  & 53.8\%  & 52.3\%  & 49.1\%  & 62.8\%  & 62.4\%  & 65.3\%  & 53.0\% \\
        \textsc{Gemini 1.5 Flash 002} & 61.5\% & 61.7\% & 53.3\% & 52.7\% & 50.0\% & 69.2\% & 67.1\% & 69.8\% & 51.8\% \\
        \textsc{Gemini 1.5 Pro} & \textbf{68.9\%} & \textbf{67.8\%} & \textbf{67.6\%} & \textbf{66.2\%} & \textbf{65.2\%} & \textbf{71.7\%} & \textbf{71.1\%} & 75.2\% & \textbf{68.0\%} \\
        \textsc{GPT 4 Turbo} & NA & NA & NA & NA & NA & NA & NA & \textbf{77.9\%} & NA \\
        \bottomrule
    \end{tabular}
    \caption{Accuracies for the tested models in the general setting. We test a broad range of captioning and reasoning models. The 001 and 002 following the Gemini flash models indicate the version tested; 002 was the newest at the time of testing.}
    \label{tab: general results full}
\end{table*}

\subsection{Error Analysis}
We perform an error analysis with 33 of the best models. For the open-sourced model, we take from the general benchmark the reasoning models \textsc{Qwen 2.5}, \textsc{NVM Llama3 70B}, and \textsc{Meta Llama 3.1 70B} with all the \textsc{ViPLava} and \textsc{BLIP2} captioning models. And from the summarization experiment, we take the \textsc{Qwen 2.5} and \textsc{Meta Llama 3.1 70B} reasoning models and pair them with the \textsc{ViPLava 7B}, \textsc{ViPLava 13B} and \textsc{BLIP2 Flan T5 XXL} captioning models after summarization by \textsc{NVM Llama3 51B} and \textsc{Qwen 2.5}. 
For the closed-sourced models, we take all \textsc{Gemini 1.5 Pro} and \textsc{GPT 4 Turbo} models from the general benchmark.

This gives 27 open-sourced models and 6 closed-sourced models. We chose these because they were generally the best-performing models, each with accuracy over or around the minimum performance criterion of 71\%.
\begin{figure}[t]
    \centering
    \begin{subfigure}{0.32\linewidth}
        \centering
        \includegraphics[width=\linewidth]{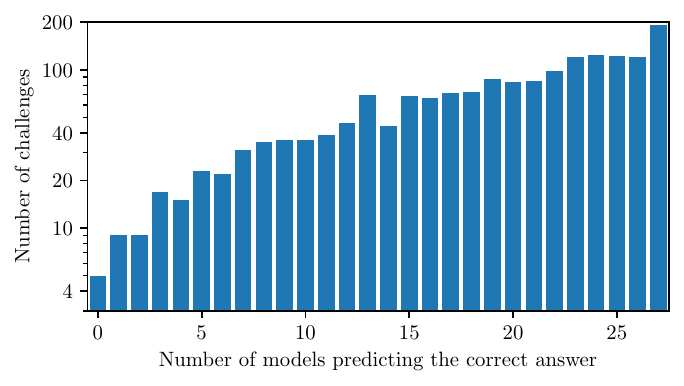}
        \caption{Open-sourced models.}
        \label{fig: correct prediction (open)}
    \end{subfigure}~
    \begin{subfigure}{0.32\linewidth}
        \centering
        \includegraphics[width=\linewidth]{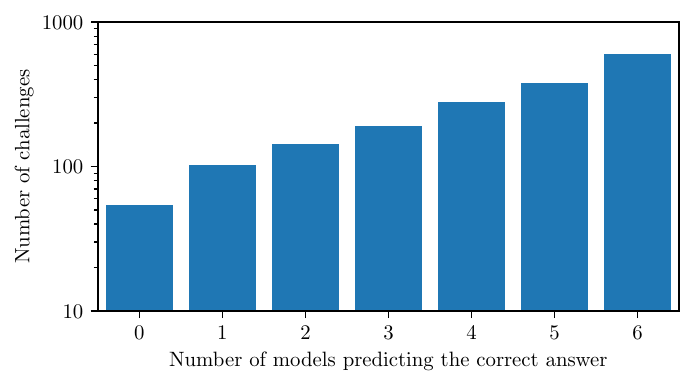}
        \caption{Closed-sourced models.}
        \label{fig: correct prediction (closed)}
    \end{subfigure}~
    \begin{subfigure}{0.32\linewidth}
        \centering
        \includegraphics[width=\linewidth]{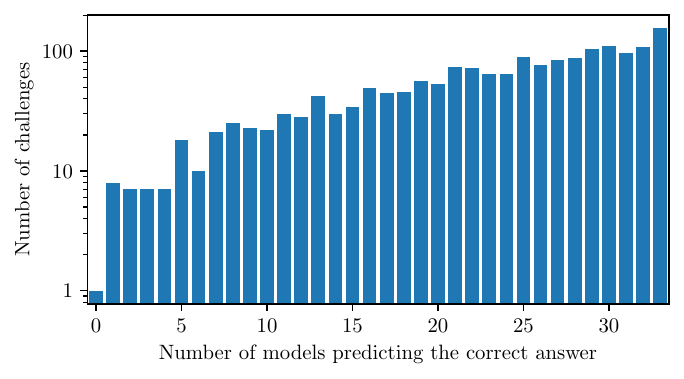}
        \caption{All models.}
        \label{fig: correct prediction (all)}
    \end{subfigure}
    \caption{Distribution of how many flips a certain number of models correctly labels. For instance, the 0 column (the leftmost column) indicates how many flips all considered models incorrectly labels. The three subfigures show the distribution for the three collections of top-performing models: only open-sourced models, only closed-sourced models, and all models. Two examples of challenges all open-sourced models fail on are shown in \Cref{fig: example of misclassified flips}.}
    \label{fig: correct predictions}
\end{figure}
We analyze the models on $\approx$1700 flip challenges from the validation data.

To analyze the performance of the best models, we compare the distributions of correct predictions across the three groups of models. \Cref{fig: correct predictions} shows the distributions of challenges correctly labeled by varying numbers of models.

\begin{figure}[t]
    \centering
    \begin{subfigure}{0.48\columnwidth}
        \centering
        \includegraphics[width=0.85\linewidth]{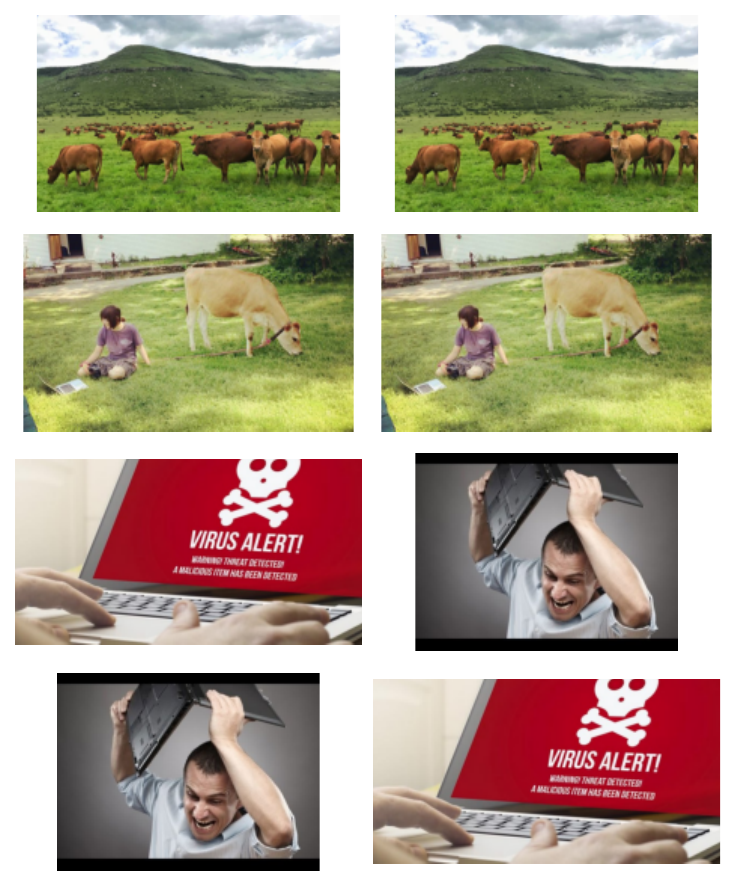} 
        \caption{First example. The correct answer is the left stack (strong consensus with 9-0 votes).}
        \label{fig: first error example}
    \end{subfigure}~
    \hfill
    \begin{subfigure}{0.48\columnwidth}
        \centering
        \includegraphics[width=0.85\linewidth]{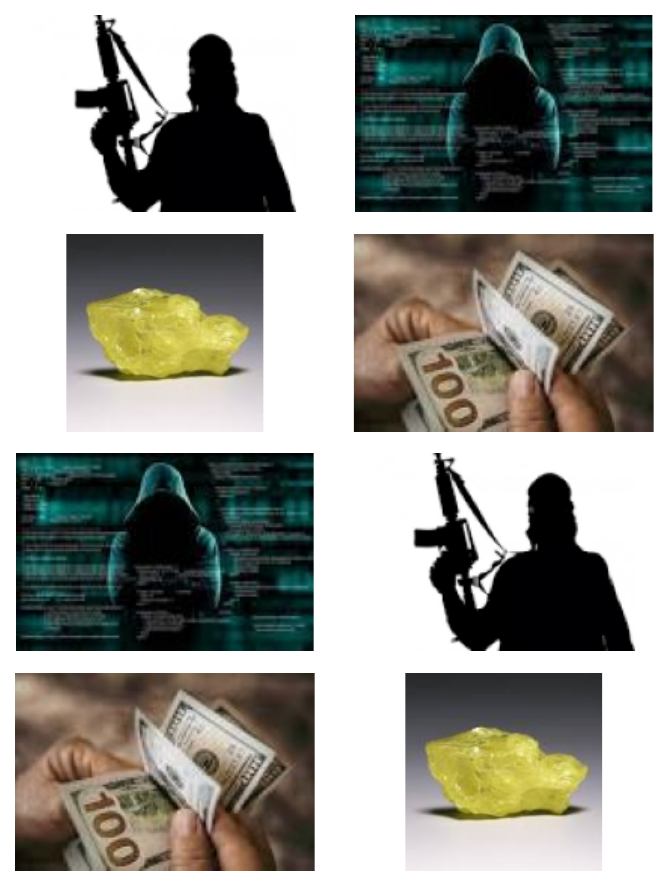} 
        \caption{Second example. The correct answer is the left stack (strong consensus with 7-2 votes).}
        \label{fig: second error example}
    \end{subfigure}
    \caption{Examples of two flip challenges where all the best open-source models fail to predict the correct answer.}
    \label{fig: example of misclassified flips}
\end{figure}

\Cref{fig: correct prediction (open)} shows the distribution for open-sourced models, where some challenges are correctly labeled by only a subset of the models, with a gradual increase as more models provide the correct answers. We show in \Cref{fig: example of misclassified flips} 2 of the challenges where all open-sourced models fail.

\Cref{fig: correct prediction (closed)} focuses on closed-sourced models, demonstrating a different trend. While the number of models is smaller, the closed-sourced models generally perform better, as evidenced by the higher number of challenges correctly labeled by more models. This may reflect the specialized training or optimization techniques applied to closed-sourced models.

\Cref{fig: correct prediction (all)} aggregates the performance of all models, combining the strengths of both open- and closed-sourced models. The broader distribution in this figure reflects the overall variability in model performance and emphasizes the need for a diverse set of models to handle challenging inputs effectively.

\begin{figure}[t]
    \centering
    \includegraphics[width=0.7\linewidth]{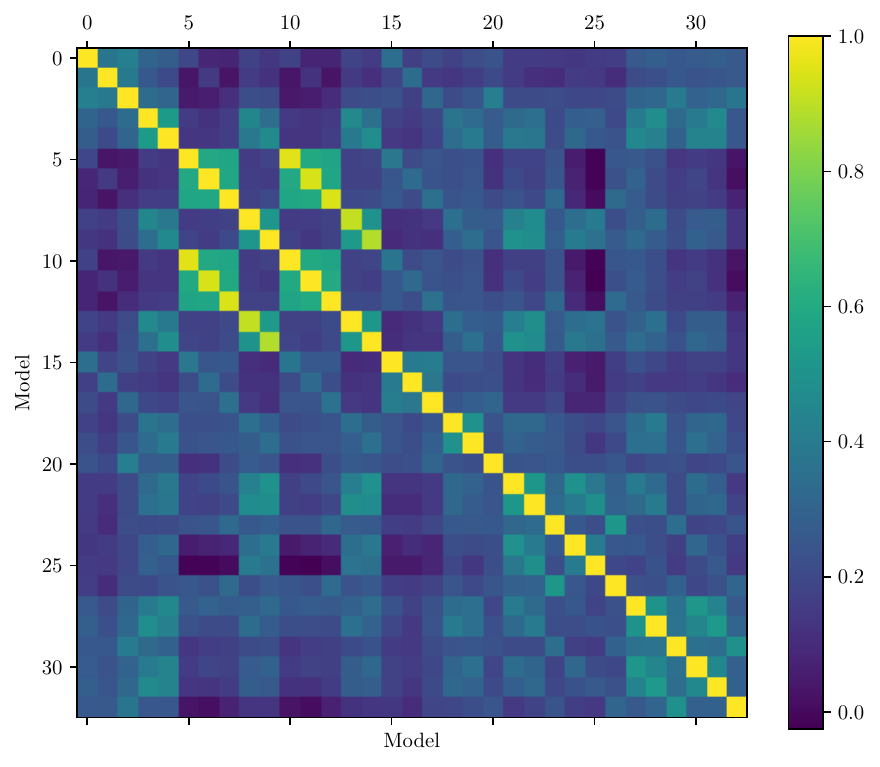}
    \caption{Correlation between model predictions for all 33 models we consider. The first 27 are the open-sourced models, while the last 6 are the closed-sourced models. }
    \label{fig: model correlation}
\end{figure}

Finally, we also measure the correlation between model prediction; cf. \Cref{fig: model correlation}. We see that the models are generally only weakly correlated. Since the models generally have accuracies around 70\% then we expect some correlation. However, we see a relatively high correlation between, e.g., the models at index 5 and 10. This corresponds to \textsc{NVM Llama3 70B} and \textsc{Meta Llama 3.1 70B} both using the same captioning model. This implies that these models are generally very similar, which is to be expected. 

\subsection{Ensemble Models}
We work with the same models as in the error analysis in the previous section. For these experiments, we consider three sets: only open-sourced models, only closed-sourced models, and both open-sourced and closed-sourced models. 
We refer the reader to \Cref{tab: general results full} with most of the considered models' performance. Overall, the best open-source closed-source models have accuracies of 74.8\% and 77.9\%, respectively. 

We present all the results of the ensembles in \Cref{tab: ensembles}.

\paragraph{Majority Voting}

We first perform a simple ensemble through majority voting, in which we consider all models of the relevant group and aggregate their predictions. This gives slightly better predictions for the ``open-source models'' and ``all models'' ensembles while the closed-sourced results are much worse (6\%-points lower). 

\paragraph{Majority Voting Using Subsets\\}

We then try to use subsets of our models to give better results with majority voting. For this, we test all subsets of size 3 and size 5 and report the highest accuracy we observe. Overall, all ensembles do better than when using a simple majority vote, and the closed-source ensemble now sees improvements over using a single model.

\paragraph{Random Forest and Logistic Regression\\}

Next, we split the data into two sets of equal size: one for training and one for testing--the presented results are on the latter. We then train a random forest and a logistic regression model on the predictions from our up to 33 models. 

Notably, we here see that the logistic regression model has an accuracy of 82.1\% using only open-sourced models. 

\paragraph{Logistic Regression on Subsets\\}

As we saw in \Cref{fig: model correlation}, there is a slight correlation between some models. We could try to regularize the fitting to bypass this, but we employ a more straightforward method here: we take random subsets of size 15. Since we no longer can compute all subsets of size 15 (33 choose 15 is 1,037,158,320, which would take days and be a matter of overfitting), we take 1,000 random samples of size 15 from both the ``open-source models'' set and the ``all models'' set. We then fit a logistic regression model for each sample as before but only using the 15 models. The result slightly improves the ensembles, resulting in the ``all models'' ensemble getting an accuracy $>85\%$.

\paragraph{Final Open-soucre Ensemble}
The best open-source ensemble consists of 15 models, 7 summarization models, and 8 from the general benchmark shown in \Cref{tab: best open source ensemble}.

\begin{table}[t]
    \centering
    \scriptsize
    \begin{tabular}{c|c}
        \toprule
        \multicolumn{2}{c}{Majority voting} \\
        \midrule
        Open-sourced models & 77.6\% \\
        Closed-sourced models & 71.9\% \\
        All models & 79.3\% \\
        \midrule[0.08em]
        \multicolumn{2}{c}{Best subset of size 3} \\
        \midrule
        Open-sourced models & 78.2\% \\
        Closed-sourced models & 78.1\% \\
        All models & 80.1\% \\
        \midrule[0.08em]
        \multicolumn{2}{c}{Best subset of size 5} \\
        \midrule
        Open-sourced models & 80.1\% \\
        Closed-sourced models & 78.5\% \\
        All models & 81.2\% \\
        \midrule[0.08em]
        \multicolumn{2}{c}{Random forest} \\
        \midrule
        Open-sourced models & 80.6\% \\
        Closed-sourced models & 77.4\% \\
        All models & 81.4\% \\
        \midrule[0.08em]
        \multicolumn{2}{c}{Logistic regression} \\
        \midrule
        Open-sourced models & 82.1\% \\
        Closed-sourced models & 79.6\% \\
        All models & 84.1\% \\
        \midrule[0.08em]
        \multicolumn{2}{c}{Logistic regression (best of 1k subsets of size 15)} \\
        \midrule
        Open-sourced models & 83.3\% \\
        Closed-sourced models & NA \\
        All models & 85.2\% \\
        \bottomrule
    \end{tabular}
    \caption{Performance of the various ensembles. The best open-source and closed-source models have accuracies of 74.8\% and 77.9\%, respectively.}
    \label{tab: ensembles}
    \vspace{-0.3cm}
\end{table}

\begin{table}[t]
    \centering
    \scriptsize
    \begin{tabular}{c|c|c}
        \toprule
        Reasoning & Captioning & Summarization \\
        \midrule
        \multirow{2}{*}{\textsc{Qwen 2.5}} & \textsc{BLIP2 2.7B} & \multirow{8}{*}{NA} \\
         & \textsc{BLIP2 Flan T5 XXL} & \\
        \multirow{3}{*}{\textsc{Meta Llama 3.1 70B}} & \textsc{BLIP2 2.7B} & \\
         & \textsc{BLIP2 6.7B COCO} & \\
         & \textsc{BLIP2 Flan T5 XXL} & \\
        \multirow{3}{*}{\textsc{NVM Llama3 70B}} & \textsc{BLIP2 6.7B COCO} & \\
         & \textsc{ViPLlava 7B} & \\
         & \textsc{ViPLlava 13B} & \\
        \midrule
        \textsc{Qwen 2.5} & \textsc{BLIP2 Flan T5 XXL} & \multirow{3}{*}{\textsc{Qwen 2.5}} \\
        \multirow{2}{*}{\textsc{Meta Llama 3.1 70B 2.5}} & \textsc{ViPLlava 7B} &  \\
         & \textsc{ViPLlava 13B} &  \\
        \multirow{2}{*}{\textsc{Qwen 2.5}} & \textsc{BLIP2 Flan T5 XXL} & \multirow{4}{*}{\textsc{NVM Llama3 51B}} \\
         & \textsc{ViPLlava 13B} & \\
        \multirow{2}{*}{\textsc{Meta Llama 3.1 70B}} & \textsc{ViPLlava 7B} & \\
         & \textsc{ViPLlava 13B} & \\
        \bottomrule
    \end{tabular}
    \caption{The models in the best ensemble of open-source models}
    \label{tab: best open source ensemble}
    \vspace{-0.4cm}
\end{table}



\end{document}